\pdfoutput=1

\documentclass[11pt]{article}

\usepackage[final]{acl}

\usepackage{times}
\usepackage{latexsym}

\usepackage[T1]{fontenc}

\usepackage[utf8]{inputenc}

\usepackage{microtype}

\usepackage{inconsolata}
\usepackage{graphicx}

\usepackage{algorithm}
\usepackage{algpseudocode}
\usepackage{amsmath} 
\usepackage{multicol}
\usepackage{subfigure}
\usepackage{booktabs}  
\usepackage{multirow}  
\usepackage{array}  
\usepackage{xspace}
\usepackage[marginpar]{todo}
\usepackage{xcolor}

\usepackage[normalem]{ulem}

\newcommand{\adaptbpe}{Adapt\textsc{BPE}\xspace{}}
\newcommand{\ourapproach}{Adapt\textsc{BPE}\xspace{}}
\newcommand{\CU}{\ensuremath{\text{CU}\xspace{}}}
\newcommand{\llama}{Llama\xspace{}}
\newcommand{\revision}[1]{\textcolor{black}{#1}}
\newcommand{\revisionVijini}[1]{\textcolor{black}{#1}}

\newenvironment{myquote}%
  {\list{}{\leftmargin=0.5em\rightmargin=0.5em}\item[]}%
  {\endlist}

\newcommand{\remove}[1]{} 

\title{Tokenizer Adaptation: From Generic to Specialized Tokenization}
\title{\adaptbpe: From Generic to Specialized Tokenizers}
\title{\adaptbpe: From General Purpose to Specialized Tokenizers}

\author{Vijini Liyanage \and François Yvon \\
  Sorbonne-Université, CNRS, ISIR, Paris \\
\texttt{\{pilanaliyanage,yvon\}@isir.upmc.fr}
  }

\begin{document}

\maketitle

\begin{abstract}
Subword tokenization methods, such as Byte-Pair Encoding (BPE), significantly impact the performance and efficiency of large language models (LLMs). The standard approach involves training a general-purpose tokenizer that uniformly processes all textual data during both training and inference. However, the use of a generic set of tokens can incur inefficiencies when applying the model to specific domains or languages. To address this limitation, we propose a post-training adaptation strategy that selectively replaces low-utility tokens with more relevant ones based on their frequency in an adaptation corpus. Our algorithm identifies the token inventory that most effectively encodes the adaptation corpus for a given target vocabulary size. Extensive experiments on generation and classification tasks across multiple languages demonstrate that our adapted tokenizers compress test corpora more effectively than baselines using the same vocabulary size. This method serves as a lightweight adaptation mechanism, akin to a vocabulary fine-tuning process, enabling optimized tokenization for specific domains or tasks. Our code and data are available at \url{https://github.com/vijini/Adapt-BPE.git}.

\end{abstract}


\section{Introduction} \label{sec:introduction}

Subword tokenization has become a standard component in modern large language models (LLMs) such as GPT, BLOOM, \llama, Gemma and Qwen  \citep{brown-2020-language,lescao-etal-2022-bloom,touvron-etal-2023-llama,mesnard-etal-2024-gemma,bai-2023-qwen}. Techniques like Byte-Pair Encoding (BPE) \citep{gage-1994-anewalgorithm,sennrich-etal-2016-neural} and SentecePiece \citep{kudo-2018-subword} represent rare and unknown words by decomposing them into frequently occurring subwords. These methods strike a balance between vocabulary size and coverage, enabling models to generalize across diverse inputs.

Despite their widespread adoption, most tokenizers operate statically. This means that once a BPE tokenizer is trained, its merge rules remain fixed throughout the model's lifecycle. This static nature can lead to suboptimal performance when the distribution of the test corpus differs from the training distribution \citep{ovadia-etal-2019-can}. In fact, learning a BPE vocabulary is no different from other learning algorithms and is prone to overfitting, hence the need include regularization mechanisms \citep{provilkov-etal-2020-bpe} or to adapt to the test conditions \citep{sachidananda-etal-2021-efficient}. Domain-specific corpora may benefit from different tokenization strategies that better capture local frequency statistics and require supplementary tokens \citep{liu-etal-2023-task,liu-etal-2024-gold}. Likewise, adapting a large multilingual model to a restricted set of languages may cause many tokens (and their associated parameters) to become useless. Adapting a tokenizer offers two potential benefits: (a) it reduces the number of units and associated model parameters that need to be managed; (b) it results in shorter tokenized texts, thereby reducing inference costs.

In this paper, we propose a lightweight and practical approach to post-hoc tokenizer adaptation. Operating under a fixed merge budget, our algorithm optimizes tokenization by replacing low-utility merges with alternatives that offer greater compression benefits for test domain(s) and language(s). This strategy, similar to model fine-tuning but operating on vocabularies, computes non-canonical tokenizations that can readily be used \emph{without updating the model weights}.
In a nutshell, our algorithm starts by applying the first $N$ merges of a pretrained BPE tokenizer. It then iteratively refines the vocabulary by replacing low-frequency tokens with high-frequency alternatives identified on a development set. This process \emph{improves compression utility}, defined as the reduction in token count, while remaining fully compatible with the associated LLM.

Unlike prior work that trains augmented domain-specific tokenizers or exploits dynamic tokenization at runtime \citep{gee-etal-2022-fast,gee-etal-2023-multi,da-dalt-etal-2024-flor,geng-etal-2025-zipzip}, our approach emphasizes a cost-benefit analysis that is particularly relevant for multilingual LLMs. These models often support vocabularies spanning thousands of tokens to accommodate as many scripts and languages as possible. However, large coverage introduces inefficiencies when processing language-specific or domain-specific corpora. By selectively refining the vocabulary, our method vastly reduces the number of active tokens, with minimal impact on performance. This reduction can yield smaller parameter sets, shorter tokenized sequences and better inference times\done\todo{Have we shown this ?}. As it does not require any change in the model weights, it offers a practical, plug-and-play solution for downstream deployment—especially in constrained or high-throughput environments.

We showcase our approach across test corpora for several LLMs, languages, domains and tasks and demonstrate its superiority over baselines operating with the same vocabulary size. Our contributions can be summarized as follows (a) we introduce a refinement algorithm that operates on BPE merge lists and improves compression utility for test corpora under a fixed merge budget; (b) we run experiments showing the effectiveness of this method in multiple monolingual and bilingual settings.


\section{Related Work} \label{sec:related}
\paragraph{Subword Tokenization.}
Subword tokenization is foundational in modern NLP. BPE \citep{gage-1994-anewalgorithm,sennrich-etal-2016-neural} is widely used in contemporary LLMs: starting with a character (or byte-based) segmention, it learns a finite set of subword units by recursively merging frequent bigrams into new units, balancing vocabulary compactness and generalization. Alternatives include Unigram LM \citep{kudo-2018-subword} and WordPiece \citep{wu-2016-google}.


\paragraph{BPE Variants.}
Extensions of BPE improve generalization or efficiency by amending the token set during learning, post-learning, or inference:
\begin{itemize}\setlength\itemsep{0pt}
\item \textbf{BPE-Dropout} \citep{provilkov-etal-2020-bpe} samples merge paths stochastically during training to improve robustness; \citep{zheng-etal-2025-broken} samples random, \emph{non-canonical} tokenizations, during inference, and observes a moderate loss of performance for fine-tuned models;
\item \textbf{BPE Trimming} \citep{cognetta-etal-2024-analysis}, \textbf{Picky BPE} \citep{chizhov-etal-2024-bpe} and \textbf{Scaffold-BPE} \citep{lian-etal-2024-scaffold} attempt to solve a defect of the BPE algorithm, which tends to produce intermediate tokens (``junk'' tokens) \citep{bostrom-durrett-2020-byte,li-etal-2024-glitch} that end up with a low frequency on the training corpus. In these variants, low-frequency tokens are removed (``trimmed'') and also possibly replaced by larger units, based on bigram counts computed on the training corpus; 
\item Another well-known defect of BPE is that it generates tokens that inconsistent with morphologically motivated segmentations \citep{vania-lopez-2017-characters,hou-etal-2023-effects,mager-etal-2022-bpe}; \textbf{BPE Knockout} \citep{bauwens-delobelle-2024-bpe} revises post-hoc the set of merge operations so as to improve the compatibility of tokens with linguistic morphemes.
\end{itemize}
Our approach also aims to replace low-utility tokens in the pretrained vocabulary; our goal is however different as we develop plug-and-play tokenization adapters that are optimized post-hoc for specific domains and languages; furthermore, it relies on the sole statistics of the adaptation corpus and does not require access to the pre-training data.

\paragraph{Vocabulary Adaptation.}
Multilingual tokenizer adaptation \cite{sachidananda-etal-2021-efficient,feng-etal-2024-adapting} has primarily been studied in a scenario where new languages need to be added to an existing language model. This typically implies adding new units, to cover more scripts \citep{lin-2024-mala,imanigooghari-etal-2023-glot500,lu-2024-llamax}, then learn the corresponding embeddings via continued pretraining. In this scenario, finding the right initialization for these new parameters improves their estimation, especially in when the adaptation data is scarce \citep{minixhofer-etal-2022-wechsel,dobler-de-melo-2023-focus,remy-etal-2024-transtokenization,singh-etal-2025-energizar}. By contrast, we consider a scenario where one want to \emph{reduce} the language coverage of a pre-trained, so as to adapt to a restricted set of texts.  

\paragraph{Vocabulary Refinement and Efficiency.}
Our approach reverses the scaling trend by identifying an effective subset of merges under a fixed budget. Unlike prior work~\cite{sachidananda-etal-2021-efficient,ushio-etal-2023-efficient,lian-etal-2024-scaffold,cognetta-etal-2024-analysis}, we do not require training data, retraining, or vocabulary heuristics. We operate directly on standard \texttt{tokenizer.json} files, using corpus-level statistics on adaptation data. Compared to \citep{bogoychev-etal-2024-ups}, which also uses heuristic vocabulary trimming in machine translation, our approach is more sound, as well as more effective.

\paragraph{Inference-Time Benefits.}
Test-time refinement yields practical efficiency: unused tokens can be pruned from the output projection matrix, reducing memory and compute costs. It also reduces off-target tokenization, improving compression and lowering the number of decoding steps—especially useful in domain-specific or low-latency settings.

\section{Methodology} \label{sec:method}

\subsection{BPE Tokenizers Adaptation}

Byte Pair Encoding (BPE) tokenizers are trained on very large corpora prior to model training. They provide an effective way to represent any running text as a sequence of tokens from a finite vocabulary, that is then used to estimate the model parameters. As they are associated with the model parameterization, tokenizers are usually kept fixed during all subsequent steps of the model lifecycle (supervised fine-tuning, adaptation, alignment, exploitation, etc). When specializing an LLM to a restricted number of languages or domains, though, the tokenization procedure may deliver suboptimal segmentations; it may also cause to load a lot of useless model parameters in memory. Our method addresses these shortcomings by revising \textit{post hoc} the set of BPE merges on an adaptation corpus.

We assume access to a BPE tokenizer (e.g., \llama{} tokenizer), from which we extract the full list of $M$ training merges. Given a target vocabulary size of $N$, we initially apply the first $N$ merges and compute the resulting token frequencies on the adaptation corpus. We then iteratively revise this merge set by replacing low-utility merges with more frequent and beneficial merges from the remaining set of merges, aiming to reduce the overall length of the adaptation corpus in a greedy fashion.

This process is applied during a post-training step: it does not require to add new tokens to the vocabulary, nor does it require access to the original training data either. As it delivers a merge list that is fully compatible with the associated LLM, it can be used without any change in the model parameter. This procedure is formalized in section~\ref{ssec:bpe-adapt}. 

\subsection{Concepts} 
Before presenting our method, we recall the main concepts of BPE-based tokenizers \citep{zouhar-etal-2023-formal,berglund-vandermerwe-2023-formalizing} and introduce useful notations. We denote $\Sigma$ a base set of symbols and $\Sigma^*$ its Kleene closure. Learning a BPE tokenizer on some training corpus $\mathcal{C}$ yields an ordered list of merges $\boldsymbol{\mu}(\mathcal{C}) = [\mu_0, \mu_1, \dots, \mu_M]$, with $\mu_i = (x_i,y_i) \in \Sigma^* \times \Sigma^*$; $x_i$ and $y_i$ will are the \emph{parent tokens} of $\mu_i$. We use Python notations to denote subsequences (e.g. $\boldsymbol{\mu}[:i]$ for $[\mu_0, \dots \mu_i]$). The tokenization process of text $\mathcal{W} \in \Sigma^*$ first\footnote{Most implementations include a pre-tokenization step, which prevents tokens to cross word boundaries. Our presentation makes no such assumption; in our experiments, we use the same pre-tokenization as the tokenizers considered.}
  splits the text in its base symbols, then recursively applies merging rules in $\boldsymbol{\mu}(\mathcal{C})$ from $\mu_0$ to $\mu_M$, where applying the rule $\mu_i = (x_i,y_i)$ replaces all occurrences of bigram $(x_iy_i)$ with a new symbol $\mu_i$. We denote $\operatorname{apply}(\mu, \mathcal{W})$ the text resulting from applying rule $\mu$ to  $\mathcal{W}$. Texts tokenized with BPE rules $\boldsymbol{\mu}$ are sequences of tokens in $(\Sigma \cup \Gamma)^*$, where $\Gamma = \operatorname{set}(\boldsymbol{\mu})$.

A key property of a learned BPE vocabulary $\boldsymbol{\mu}(\mathcal{C})$ is \emph{properness} \cite{berglund-vandermerwe-2023-formalizing}: a merge sequence is \emph{proper} if every compound token is only created after its parents have been created. Formally, 
a merge sequence $\mu = [\mu_0, \mu_1, \dots, \mu_{N}]$ is \emph{proper} if for all $\mu_i = (x_i, y_i)$: (a) $x_i \in \Sigma \text{ or } \exists j < i | x_i = \mu_j$ ; (b) $y_k \in \Sigma \text{ or } \exists k < i | y_i = \mu_k$. Note that any prefix of a proper merge sequence remains proper.

Following \citet{lian-etal-2024-scaffold}, we extend these notions by distinguishing between \emph{actual} and \emph{virtual} (or scaffold) merges, with the following semantic: actual merges correspond to tokens that will appear in the tokenized text; virtual merges only exist to create larger units, and do not appear in the tokenized text. Given a list of merges $\boldsymbol{\mu}$ and their associated type, tokenization proceeds as follows: (a) merges in $\boldsymbol{\mu}$ are applied from first to last; (b) $\boldsymbol{\mu}$ is then processed \emph{backwards}\footnote{This ensures that virtual tokens are split while their parents are still merged.} and virtual tokens are replaced with their parents -- a procedure denoted $\operatorname{unapply}(\mu, \mathcal{W})$ below.

\subsection{The \adaptbpe{} Algorithm \label{ssec:bpe-adapt}}
\adaptbpe{} is formalized in Algorithm~\ref{alg:bpe-adapt}. It maintains the following invariant: at any stage, $\boldsymbol{\mu}_\text{A}$ is a proper sequence of exactly $N$ actual merges, containing only merges from the initial tokenizer. This is true initially (line~1). Exchanging $\mu_p$ for $\mu_q$ (lines~11-12)\footnote{In line 11, we append a new element to list $\boldsymbol{\mu}_\text{A}$; in line 12 we \emph{remove} an element from the list $\boldsymbol{\mu}_\text{R}$.} does not change the number of actual merges. Furthermore, as we only promote tokens from $\boldsymbol{\mu}_R$ when both parents exist in $\boldsymbol{\mu}_\text{A}$, this exchange also preserves the properness of the merge list. This procedure is illustrated in Appendix~\ref{ssec:illustration}.

As we rely on empirical counts the development corpus, the frequency computations can be unreliable, especially for the rarer events. The end condition (line~9) can be adapted to this uncertainty, for instance by ensuring the frequencies of the incoming and deleted unit differ by some margin.

The greedy merge procedure ensures that each replacement results in a net decrease of the total corpus size. Like for the original BPE algorithm, the complexity is dominated by the cost to compute and update sorted frequency lists, which can be done effectively using appropriate data structures.\footnote{\url{https://guillaume-be.github.io/2021-09-16/byte_pair_encoding} or \url{https://github.com/marta1994/efficient_bpe_explanation}.}

\begin{algorithm}[tb]
\caption{BPE adaptation} 
\label{alg:bpe-adapt}
\begin{algorithmic}[1]
\Require Merge list $\boldsymbol{\mu} = [\mu_0, \dots, \mu_M]$, adaptation corpus $\mathcal{A}$, merge budget $N$
\State $\boldsymbol{\mu}_{\text{A}} \gets [:\mu_{N-1}]$, $\boldsymbol{\mu}_{\text{R}} \gets [\mu_N:\mu_M]$
\State $\mathcal{A}^t \gets \operatorname{apply}(\boldsymbol{\mu}_{\text{A}},\mathcal{A})$
\State Compute unigram frequencies: \\ $\quad \operatorname{Freq}(\mu, \mathcal{A}^t), \forall \mu \in \boldsymbol{\mu}_{\text{A}}$
\State Compute bigram frequencies: \\ $\quad \operatorname{Freq}(\mu\mu', \mathcal{A}^t), \forall (\mu, \mu') \in \boldsymbol{\mu}_{\text{A}} \times \boldsymbol{\mu}_{\text{A}} $
\State $\mu_p \gets \arg \min_{\mu \in \boldsymbol{\mu}_\text{A}} \operatorname{Freq}(\mu, \mathcal{A}^t)$ 
\State $\mu_q \gets \arg \max_{\mu = (x,y) \in \boldsymbol{\mu}_\text{R}} \operatorname{Freq}(xy, \mathcal{A}^t)$ 
\While{$\operatorname{Freq}(\mu_p, \mathcal{A}^t) < \operatorname{Freq}(\mu_q, \mathcal{A}^t)$}
    \State Make $\mu_p$ a virtual merge
    \State $\mathcal{A}^t \gets \operatorname{unapply}(\mu_p, \mathcal{A}^t)$
    \State $\mathcal{A}^t \gets \operatorname{apply}(\mu_q, \mathcal{A}^t)$
    \State $\boldsymbol{\mu}_{\text{A}} \gets \boldsymbol{\mu}_{\text{A}} + [\mu_q]$
    \State $\boldsymbol{\mu}_{\text{R}} \gets \boldsymbol{\mu}_{\text{R}} - [\mu_q]$
    \State Update unigram and bigram frequencies
    \State $\mu_p \gets \arg \min_{\mu \in \boldsymbol{\mu}_\text{A}} \operatorname{Freq}(\mu, \mathcal{A}^t)$ 
\State $\mu_q \gets \arg \max_{\mu = (x,y) \in \boldsymbol{\mu}_\text{R}} \operatorname{Freq}(xy, \mathcal{A}^t)$ 
\EndWhile

\State \Return $\boldsymbol{\mu}_\text{A}$
\end{algorithmic}
\end{algorithm}

\section{Experimental Setup \label{sec:experiments}}

\subsection{Overview}

Our experiments evaluate the following  scenarios. \paragraph{Monolingual adaptation:} In this setup, we adapt an LLM \emph{to process text in only one language, possibly in a restricted domain}. For each language, we collect adaptation data that are used to trim the vocabulary. \revision{We then compare the original tokenizer with adapted versions on generation and classification tasks.}

\paragraph{Bilingual adaptation} In this setup, we adapt the tokenizer to perform machine translation (MT) in \emph{one single language pair}. We optimize the merge list using the two adaptation corpora for these languages, and evaluate the results using MT metrics.

\subsection{Data \label{ssec:data}}

We use \revision{several} data sets in our experiments: Wikipedia, PubMed and \revision{SIB} for monolingual evaluations, FLORES and EMEA for bilingual ones.

\paragraph{Wikipedia.}

We collect 100 articles from the English, French, German, Manipuri, Occitan, Tamil, Spanish, and Swahili editions of Wikipedia. Articles are retrieved from curated “featured article” lists\footnote{\url{https://en.wikipedia.org/wiki/Wikipedia:Featured_articles}} to ensure content quality.
For English, French, German, and Spanish, we restrict our collection to articles promoted between 2024 and 2025. This time window reduces the risk of overlap with the pretraining corpora of LLMs such as \llama-3 and BLOOM. For the low-resource languages (Manipuri, Occitan, Swahili and Tamil), no such time restriction applies. Even then, due to the scarcity of content in Manipuri, we were only able to collect 97~articles for that language. All articles were parsed, cleaned, and segmented into sentences using a custom extraction pipeline. Statistics are in Table~\ref{tab:corpus-stats}.
The resulting data is randomly split into 50\% development and 50\% test sets. The development set is used to compute the optimal list of BPE merges for each language. The test set is used to evaluate the resulting tokenizers in terms of compression utility and perplexity. 

\begin{table}[]
  \resizebox{\columnwidth}{!}{ 
    \begin{tabular}{|l|r|l|r|} \hline
      Language & \multicolumn{1}{c|}{Length} & Language & \multicolumn{1}{c|}{Length}\\ \hline
      English (eng)   & 1943684 &  French  (fra)    & 2725460\\ 
      German  (deu)  & 614542 &  Spanish (spa)   & 2443181\\
      Occitan (oci)    & 693123 & Manipuri (mni) & 207892 \\ 
      Swahili (swa)    & \revision{176289} &  Tamil (tam)   & \revision{375334}\\ \hline
    \end{tabular}
  }
  \caption{Corpus Extracted from Wikipedia. Length is a number of bytes.}
  \label{tab:corpus-stats}
\end{table}

\paragraph{SIB.} \revision{The SIB-200 multilingual benchmark is designed to evaluate topic classification across over 200 languages \cite{adelani-etal-2024-sib}. It consists of articles annotated with eight topics: \emph{science}, \emph{technology}, \emph{travel}, \emph{politics}, \emph{sports}, \emph{health}, \emph{entertainment}, and \emph{geography}. We use it for zero-shot topic classification experiments in the same eight languages.}
\revision{In our experiments, we rely on the official train–test splits for SIB.\footnote{\url{https://huggingface.co/datasets/Davlan/sib200}} The English training set is used to fine-tune the \llama{} model; evaluations are performed with the official test sets.}

\paragraph{FLORES.}

FLORES \cite{nllb-2022} is a multilingual benchmark consisting of professionally translated sentence pairs across 200 languages. We focus on two translation tasks, \textbf{English–French} and \textbf{English–Occitan}, respectively illustrating a high-resource and a low-resource language pair.

For all the experiments on Wikipedia, SIB and FLORES, \adaptbpe{} relies on the corresponding Wikipedia development sets. We additionally consider specialized monolingual and bilingual corpora in the Medical domain, to showcase a setting where adaptation simultaneously targets a specialized domain and a subset of languages.

\paragraph{PubMed and EMEA}\done\todo{write this}
\revisionVijini{PubMed\footnote{\url{https://pubmed.ncbi.nlm.nih.gov}}
 is a large English biomedical corpus containing scientific abstracts and articles, while EMEA \citep{tiedemann-2012-parallel} consists of English–French parallel texts from the European Medicines Agency. We use a recent (2025) snapshot of PubMed and the latest version of EMEA to extract monolingual English data (PubMed) and bilingual English–French data (EMEA) for domain adaptation experiments.}

\subsection{Models and their Tokenizer \label{ssec:models}}

Our experiments use three tokenizers and models.\footnote{Appendix~\ref{ssec:app:cupp} additionally reports results obtained with Qwen-3 \citep{bai-2023-qwen}}.

\textbf{BLOOM:} Based on a byte level BPE tokenizer, BLOOM \citep{lescao-etal-2022-bloom} uses a vocabulary of 250,880 tokens and supports multilingual input. It was trained on ROOTS \cite{laurencon-etal-2022-roots} a multilingual corpus spanning texts in 46 languages. In our experiments, we use the smallest BLOOM model (560m parameters). 

\textbf{\llama-3:} Developed for the \llama-3 models \cite{grattafiori-etal-2024-llama3}, this BPE tokenizer has a vocabulary size of approximately 128k tokens. 
We experiment with the 8b parameter models.

\textbf{GPT-2:} GPT-2 relies on byte-level BPE with a vocabulary of about 50k tokens. This tokenizer is primarily optimized for English and exhibits little multilingual coverage. We use it to contrast multilingual tokenizers with an English-centric one.

\subsection{Baselines \label{ssec:baselines}}

Our experiments evaluate the cost-benefit trade-off of restricting the tokenizer vocabulary to a predefined size \(N\). We compare our results to baselines that use a limited vocabulary of size $N$:
\begin{enumerate}
    \item \textbf{Baseline 1 (First$_k$) - Merge-Truncated Tokenization:} We apply only the first $N$ merges from the pretrained tokenizer to the corpus, without any refinement or filtering.
    
    \item \textbf{Baseline 2 (First$_{k>0}$) - Merge-Truncated Tokenization with Data-aware Filtering:} We apply the first $N$ merges that appear at least once in the dataset, thereby excluding zero-frequency merges before tokenization.\footnote{In the terms of Section~\ref{sec:method}, this is done by making the zero-frequency tokens virtual, so as to preserve properness.}

    \item \textbf{Baseline 3 (Top$_k$) - Full Merge with Selective Unmerging:} We first apply all merges defined by the original tokenizer to the corpus. We then record the $N$ most frequent tokens. Less frequent tokens are recursively unmerged until their subparts belong to the top-$N$ list or to the base tokens. The resulting vocabulary thus matches the size constraint.
\end{enumerate}
For completeness, we also report the performance achieved with the full tokenizer vocabulary.

\subsection{Setup and Metrics \label{ssec:setup}}

\subsubsection{Monolingual Generation}
We report results using two main metrics.

\paragraph{Compression Utility} (CU) \citep{zouhar-etal-2023-formal}, is defined as the relative reduction in corpus size (in characters) after tokenization:
  \[
    \CU{} = \frac{|\mathcal{W}| - |\operatorname{apply}^*(\boldsymbol{\mu}, \mathcal{W})|}{|\mathcal{W}|}
  \]
  where $|\mathcal{W}|$ is the length of text $\mathcal{W}$, and $\operatorname{apply}^*(\boldsymbol{\mu},\mathcal{W})$ recursively applies all the merges in $\boldsymbol{\mu}$ in their appearance order to text $\mathcal{W}$. \CU{} measures the relative reduction in corpus size after applying the tokenizer merges and has been showed to correlate well with performance on downstream tasks \citep{goldman-etal-2024-unpacking}. Compression utility is also directly related to the \emph{tokenizer fertility}\footnote{Fertility is the average number of tokens per word, equal to $\frac{\text{CU} -|\mathcal{W}|}{L_W}$, with $L_W$ the length in words.} and to the length of the tokenized files; it thus has a direct impact on computation costs.
  
\paragraph{Perplexity} \citep{cover-thomas-1991-elements} quantifies the model's uncertainty in predicting the next token. 
We report \emph{word-level perplexity} computed with \texttt{lm-evaluation-harness} \citep{eval-harness},\footnote{Version: \texttt{0.4.8}, commit: \texttt{\detokenize{92d6139}}} enabling cross-tokenizer comparisons.

\subsubsection{Zero-shot Cross-lingual Classification}

\revision{Following \citet{sanh-2022-multitask} and \citet{adelani-etal-2024-sib}, we adopt a simple template for zero-shot cross-lingual classification. Each input sentence is wrapped in the following prompt:}
\begin{quote}
\texttt{``Is this a piece of news regarding \{science/ technology, travel, politics, sports, health, entertainment, or geography\}? \{INPUT\}''}
\end{quote}

\revision{The task consists of assigning each input sentence to one of the seven predefined categories. We report classification accuracy on the SIB test partition as the main metric.} \revision{We first fine-tune \llama{}-3.1-8b with the English training data, then contrast two experimental settings:}
\done\todo{How do you get the model's answer? Thanks to FT, the model always generate as prefix }

\begin{enumerate}{}
\item Using the fine-tuned \llama{} model with the full tokenizer to perform topic classification;
\item Using the fine-tuned \llama{} model with a language-adapted BPE tokenizer, considering a restricted vocabulary of $N$ units.\footnote{Details regarding SIB are in appendix~\ref{ssec:app:sib}.}
\end{enumerate}

\subsubsection{Machine translation}

We use a 5-shot setup for translation. The first five sentence pairs in FLORES-200 test file are selected as in-context samples (and excluded from the BLEU computation) in the following template:
\begin{quote}
\texttt{``Translate the following French sentences to English:}\\
\texttt{French: [source text]}\\
\texttt{English:[target text]}\\
\texttt{...}\\
\texttt{French: [input text]}\\
\texttt{English:''}
\end{quote}
The last source sentence in the prompt will be translated. We evaluate translation quality using BLEU \citep{papineni-etal-2002-bleu} computed with \texttt{sacreBLEU} \citep{post-2018-call}.\footnote{Signature: \texttt{nrefs:1|case:mixed|eff:no|tok:13a| \linebreak smooth:exp|version:2.5.1}} We use greedy decoding with maximum generation limit of 128 tokens for all models.

Translation differs from the other tasks because we need to \emph{actually generate texts with a reduced vocabulary}: to prevent the decoding of tokens that have been pruned, we apply logit masking in the unembedding layer so as to make the generation of such tokens impossible.\done\todo{Explain decoding alg}    

\section{Results and Analysis \label{sec:results}}

\begin{figure*}[tbp]
\label{fig:varying-k}
  \centering
  \subfigure[French]{%
    \includegraphics[width=0.95\columnwidth]{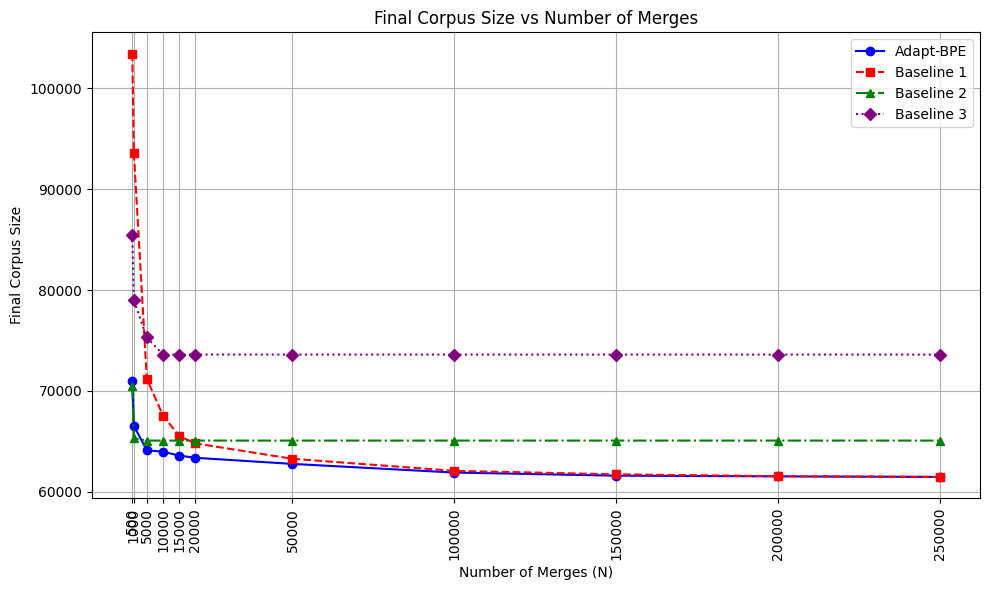} 
   \label{fig:varying-k-french}
} \quad
\subfigure[Occitan]{%
  \includegraphics[width=0.95\columnwidth]{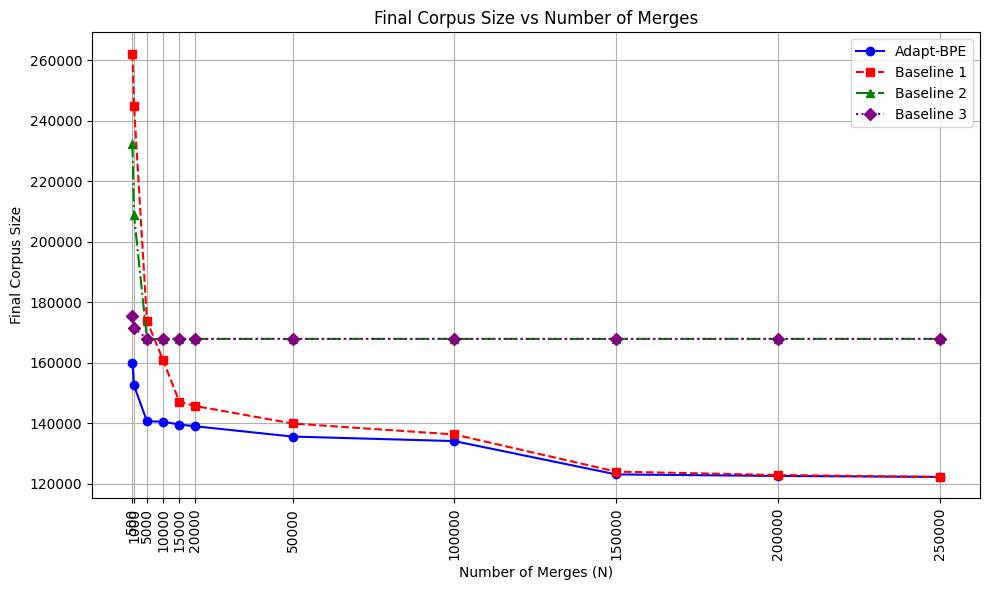} 
 \label{fig:varying-k-occitan}
}
\caption{Corpus size with increasing number of merges (BLOOM Tokenizer). }
\end{figure*}

In preliminary experiments, we study the effect of varying the number of merges on the total corpus size. We experiment with BLOOM, for which the training set is precisely documented,\footnote{\url{https://huggingface.co/bigscience/bloom\#training-data}} and consider two languages:  French, which accounts for about 13\% of the pretraining tokens, and Occitan, which is hardly represented (Occitan).\footnote{The processing of Occitan is facilitated by the large share of Romance languages 
in BLOOM's pretraining data.}

We display the learning curves on the development sets on Figures~\ref{fig:varying-k-french}, ~\ref{fig:varying-k-occitan}: each value represent the corpus size after running \adaptbpe{} (or the baselines) with a total budget of $N$.\footnote{Corpus sizes are thus monotonically decreasing.} In both cases, we observe a sharp decrease of the corpus size for the first thousands of merges for all methods, with \adaptbpe{} however always yielding shorter corpus sizes than alternative pruning methods. As the merge budget reaches 10k to 20k, the return of each additional merge gets smaller for all methods, slowly converging to the tokenization obtained with the complete tokenizer and its 250k merges.

  
For all follow-up experiments, we use a budget of 15k, which for BLOOM corresponds to a small fraction (about 6\%) of the vocabulary and the corresponding embedding parameters: for the smallest model (with 560m parameters), embeddings account for about 40\% of the parameters. The larger \llama{} (8b) only contains about 500m embeddings parameters, corresponding to about 6.5\% of the total parameter set.

\subsection{Monolingual Evaluations} 

\begin{table*}[h!]
\centering
\setlength{\tabcolsep}{4pt} 
\renewcommand{\arraystretch}{0.95} 
\resizebox{\textwidth}{!}{%
  \begin{tabular}{|l|*{4}{rrr|}}
    \toprule
    & \textbf{GPT-2 } & \textbf{BLOOM } & \textbf{\llama-3 } 
    & \textbf{GPT-2 } & \textbf{BLOOM } & \textbf{\llama-3 }
    & \textbf{GPT-2 } & \textbf{BLOOM } & \textbf{\llama-3 }
    & \textbf{GPT-2 } & \textbf{BLOOM } & \textbf{\llama-3 } \\
    \toprule
    & \multicolumn{3}{c|}{English (1.32m)} 
    & \multicolumn{3}{c|}{French (0.97m)} 
    & \multicolumn{3}{c|}{Occitan (0.34m)} 
    & \multicolumn{3}{c|}{Manipuri (0.04m)} \\
    \midrule
    Full vocab        & 0.66/09.9  & 0.66/05.9  & 0.66/04.5 
                      & 0.62/14.9 & 0.63/08.5  & 0.63/06.7 
                      & 0.59/04.8  & 0.60/05.8  & 0.66/04.0 
                      & 0.07/02.2  & 0.07/03.6  & 0.08/01.9  \\
    First$_k$         & 0.62/13.2 & 0.61/07.7  & 0.62/05.1  
                      & 0.56/17.9 & 0.59/10.6 & 0.57/08.4 
                      & 0.55/06.0  & 0.56/06.1  & 0.60/05.9 
                      & 0.03/03.2  & 0.03/03.7  & 0.03/02.3 \\
    First$_{k>0}$     & 0.63/11.3 & 0.63/07.0  & 0.64/05.0  
                      & 0.59/16.1 & 0.61/09.9  & 0.61/07.8 
                      & 0.56/05.9  & 0.58/06.5  & 0.63/05.3 
                      & 0.05/03.3  & 0.05/03.8  & 0.05/02.0 \\
    Top$_k$           & 0.63/11.8 & 0.64/06.5  & 0.64/04.8  
                      & 0.59/16.1 & 0.61/09.4  & 0.61/07.6 
                      & 0.58/05.7  & 0.57/06.1  & 0.58/04.4 
                      & 0.03/03.3  & 0.03/03.5  & 0.03/02.0 \\
    \adaptbpe         & 0.64/10.1 & 0.64/06.1 & 0.64/04.7
                      & 0.61/15.3 & 0.61/09.0 & 0.61/07.0
                      & 0.58/05.0 & 0.59/06.0 & 0.65/04.2
                      & 0.06/02.3 & 0.06/03.7 & 0.07/02.0  \\
    \bottomrule
  \end{tabular}
}
\caption{Compression utility and perplexity for $N=15$k merges on the Wikipedia test sets. The raw corpus size (before tokenization) is given in parentheses alongside each language. Each cell reports utility/perplexity.}
\label{tab:monolingual}
\end{table*}

\paragraph{Wikipedia Corpus} Table~\ref{tab:monolingual} reports both compression utility and perplexity scores for \revision{four} languages using tokenizers adapted under a merge budget of 15k.\footnote{Full results, for 8 languages, are in Appendix~\ref{ssec:app:cupp}.}
As expected, using the complete vocabulary yields the best performance across the board. Among models relying on a constraint vocabulary, \ourapproach{} consistently yields better compression utilities than the three baseline methods (see section~\ref{ssec:baselines}), resulting in compact tokenizations for all languages and tokenizers, almost closing the gap with the full vocabulary.

While the Full-Vocab tokenizer tends to achieve lower perplexity (and higher compression utility) on general-domain corpora (e.g., Wikipedia), Adapt-BPE consistently matches or exceeds performance on domain-specific corpora, such as PubMed and EMEA, as shown in Table~\ref{tab:combined_monolingual_MT}.

In particular, our method improves over the $top_{k>0}$ baseline: even though both approaches are close, \adaptbpe{} still achieves better compression scores. This is because its merging policy is based on more reliable frequency estimates performed on the adaptation data.
Using a reduced vocabulary does not seem to harm perplexity scores either: they are better or comparable than the baselines using a 15k vocabulary, demonstrating that our strategy to select merges better aligns tokenization with model expectations. The benefits of our method are especially clear for morphologically complex or low-resource languages such as Occitan and Manipuri, where other merge selections techniques seem to struggle.

\paragraph{Merge Depth Analysis}

We report in Table~\ref{tab:merge_depth_main} the index of the last selected merge for each language after refinement under a fixed 15k merge budget. This index reflects how far in the BLOOM merge list the algorithm had to go to adapt the vocabulary. 
\done\todo{add missing languages}

\begin{table}[h]
\centering
\resizebox{\columnwidth}{!}{
\setlength{\tabcolsep}{3pt}
\begin{tabular}{cccccccc}
\toprule
eng & fra & deu & spa & oci & mni & swh & tam \\
\midrule
18678 & 25784 & 22580 & 30047 & 115565 & 180908 & 84037 & 91135\\
\bottomrule
\end{tabular}
}
\caption{Index of last merge (BLOOM tokenizer).}
\label{tab:merge_depth_main}
\end{table}

High resource languages like English, French, German, and Spanish have relatively low indices, indicating that they are well covered with the frequent merges. For Occitan and Manipuri, these values are much higher, reflecting their small representation in the pretraining data.

\paragraph{Medical Domain}
\begin{table*}[htb!]
\centering
\setlength{\tabcolsep}{4pt} 
\renewcommand{\arraystretch}{0.95} 
\resizebox{0.95\textwidth}{!}{%
\begin{tabular}{|l|*{6}{r|} | r| r|}
\multicolumn{1}{c}{} & \multicolumn{6}{c}{\textbf{Compression utility / Perplexity}} & \multicolumn{2}{c}{\textbf{BLEU scores}} \\
\toprule
& \multicolumn{2}{c|}{PubMed (1.29m)} 
& \multicolumn{2}{c|}{EMEA-eng (39.48m)} 
& \multicolumn{2}{c||}{EMEA-fra (45.76m)} 
& \multicolumn{2}{c|}{EMEA eng$\rightarrow$fra} \\
\midrule
& \textbf{BLOOM} & \textbf{\llama-3} 
& \textbf{BLOOM} & \textbf{\llama-3} 
& \textbf{BLOOM} & \textbf{\llama-3} 
& \textbf{BLOOM} & \textbf{\llama-3}\\
\midrule
Full vocab       & 0.62/18.2 & 0.62/16.6 & 0.62/13.2 & 0.60/12.6 & 0.60/16.1 & 0.63/14.3 & 42.9 & 45.4 \\
First$_k$        & 0.60/18.6 & 0.59/17.8 & 0.60/14.1 & 0.59/13.3 & 0.58/16.9 & 0.61/15.9 & 40.3 & 43.8 \\
First$_{k>0}$    & 0.60/17.3 & 0.60/16.8 & 0.60/13.3 & 0.59/13.1 & 0.59/16.6 & 0.62/15.8 & 40.6 & 44.5 \\
Top$_k$          & 0.61/15.6 & 0.60/16.3 & 0.60/10.2 & 0.60/12.7 & 0.60/16.6 & 0.62/14.7 & 41.2 & 44.9 \\
\adaptbpe        & \textbf{0.62}/\textbf{11.4} & \textbf{0.65}/\textbf{12.5} & \textbf{0.64}/\textbf{6.7} & \textbf{0.60}/\textbf{8.0} & \textbf{0.60}/\textbf{12.2} & \textbf{0.63}/\textbf{11.8} & \textbf{42.9} & \textbf{45.6} \\
\bottomrule
\end{tabular}%
}
\caption{Compression utility / perplexity scores for $N=15$k merges on PubMed (English) and EMEA (English and French) test sets \textbf{(left side)}, and BLEU scores for 5-shot English–French translation on the EMEA test set \textbf{(right side)}. Corpus sizes in bytes are in parentheses.}
\label{tab:combined_monolingual_MT}
\end{table*}

Table~\ref{tab:combined_monolingual_MT} reports the performance of \adaptbpe{} on two medical corpora (PubMed, in English and EMEA, in English and French). Regarding compression, we again observe that our method does better than alternatives, and closes the gap with the full vocabulary setup. Perplexity scores show an even better trend
as we observe values that are \emph{significantly better than those obtained with the complete vocabulary}.\done\todo{normalizer ?}

\subsection{Zero-shot Text Classification \label{ssec:textclass}}

Table~\ref{tab:classification_results} presents zero-shot classification results when using either a full vocabulary in inference, or a language-adapted token set of 15k merges.\done\todo{rational for language ordering ?}
\begin{table}[t]
\centering
\resizebox{\columnwidth}{!}{%
\setlength{\tabcolsep}{2pt}
\begin{tabular}{l|ccccccccc}
\toprule
\textbf{Setting} & \textbf{eng} & \textbf{fra} & \textbf{deu} & \textbf{spa} & \textbf{oci} & \textbf{mni} & \textbf{swh}  & \textbf{tam} & \textbf{sin}\\
\midrule
FT + Full vocab   & 54.9  & 52.0 & 48.1 &  43.2 & 32.1 & 16.3 & 25.2 & 23.9 &  17.2\\
FT + \adaptbpe  & 46.5  & 48.4 & 46.2 &  41.0 & 30.2 & 16.3 & 24.9 & 23.3 & 17.1 \\
\adaptbpe + FT  & 45.3  & 48.8  & 47.5 & 42.4 & 31.2 & 16.3 & 25.4 & 23.5 & 17.2 \\
  \bottomrule
\end{tabular}
}
\caption{Zero-shot classification accuracy for fine-tuned (FT) \llama-3 models with a full or trimmed vocabulary.}
\label{tab:classification_results}
\end{table}
\revision{We observe here a large drop in performance of English, owing to the mismatch between training data (using the original tokenizer) and test data (using a reduced one): rarer tokens are helpful in topic classification and their removal hurts performance. Already for French, the impact is lessen, and progressively vanishes for Spanish and German, down to the low-resource language set, where the performance remains poor. This shows that \emph{cross-lingual transfer is robust to vocabulary pruning}, and its effect subsist even when only using  a small portion of the original vocabulary. In an additional experiment, we \emph{combine \adaptbpe{} and fine-tuning}, recovering part of the accuracy loss for all languages but English and Tamil.
}

\subsection{Machine Translation \label{ssec:machtrans}}   

Table~\ref{tab:MT} reports BLEU scores on FLORES datasets for English-French and English-Occitan translation using BLOOM and \llama-3 models (5-shot setting). We evaluate two merge budgets: 15k, corresponding to the monolingual vocabulary size of previous sections, and 30k, doubling the vocabulary for the bilingual setting.
The refined tokenizer consistently improves translation quality relative to baselines and closely matches the performance of the full model at a reduced computational cost (details in Appendix~\ref{ssec:ann:compute}). These results illustrate the potential payoffs of using \adaptbpe{}.  
\begin{table}[ht]
\centering
\resizebox{\columnwidth}{!}{%
\begin{tabular}{l|rr|rr||rr|rr}
\toprule
\textbf{Method} 
& \multicolumn{4}{c||}{\textbf{BLOOM}} 
& \multicolumn{4}{c}{\textbf{\llama-3}} \\
& \multicolumn{2}{c|}{\textbf{eng-fra}} & \multicolumn{2}{c||}{\textbf{eng-oci}} 
& \multicolumn{2}{c|}{\textbf{eng-fra}} & \multicolumn{2}{c}{\textbf{eng-oci}} \\
& 15k & 30k & 15k & 30k 
& 15k & 30k & 15k & 30k \\
\midrule
Full vocab        & 39.5 & 40.2 & 25.3 & 26.2 & 40.4 & 40.8 & 27.0 & 27.8 \\
First$_k$        &  30.3 & 34.9 & 18.2 & 20.5 & 32.0 & 36.5 & 19.0 & 21.1 \\
First$_{k>0}$    &  35.2 & 36.9 & 21.7 & 23.4 & 36.0 & 38.2 & 23.5 & 24.6 \\
Top$_k$          &  37.1 & 38.2 & 22.3 & 24.1 & 37.9 & 39.0 & 24.1 & 25.7 \\
\adaptbpe        & \textbf{38.9} & \textbf{39.1} & \textbf{24.8} & \textbf{25.9} & \textbf{39.7} & \textbf{40.2} & \textbf{26.3} & \textbf{27.2} \\
\bottomrule
\end{tabular}
}
\caption{BLEU scores on FLORES test with 5-shot MT with merge budgets of 15k and 30k.}
\label{tab:MT}
\end{table}

Table~\ref{tab:combined_monolingual_MT} (right) displays additional MT results obtained with a corpus of medical texts, where we observe a similar trend: better results for \adaptbpe{} than for baselines, matching the scores obtained with the full tokenizer, at a reduced cost. 

\section{Conclusion \label{sec:conclusion}}

This work studies the adaptation of a pretrained BPE vocabulary in contexts where a model is specialized (e.g., by fine-tuning) on a small number of domains or languages. For this, we propose \adaptbpe{}, a vocabulary refinement algorithm that does not imply any change in the model weights.

We show that trimming the token set can act as a form of regularization: by constraining the vocabulary size, it biases the model toward more generalizable subword representations. A key property of our lightweight approach is that it minimally intervenes on the pretrained tokenizer by (a) marking merges that need to be undone; (b) truncating the set of merges to a prefix. As such, the resulting tokenizer remains fully compatible with the pretrained model and its derivatives (e.g., fine-tuned versions). Furthermore, the performance observed with \adaptbpe{} are achieved prior to model adaptation; they would likely increase after finetuning on the adaptation data.\done\todo{As they would also benefit from initial tokenizers trained with a regularization.}

In our future work, we would like to develop methods that optimize the choice of the budget \(N\) dynamically, potentially varying according to the targeted task(s). Using a predefined number of merges can yield suboptimal vocabularies. Another improvement will be to better take the uncertainty of counts estimates into account, as using raw count might cause our algorithm to overtrain its token inventory. 


\section{Limitations \label{sec:limitations}}
\adaptbpe{} is a post-hoc refinement method that improves tokenizer efficiency without modifying model parameters.

However, it does not benefit from joint optimization with the model, which could unlock greater improvements. Additionally, the merge budget \(N\) is manually fixed; determining an optimal value remains an open question.

Finally, our experiments focus on a small set of multilingual BPE-based tokenizers and a limited set of models, languages and tasks, leaving generalization to other settings for future exploration.

\section*{Acknowledgments}

This work was funded by the French Agence Nationale de la Recherche (ANR) under the project
TraLaLaM (“ANR-23-IAS1-0006”). This work was also granted access to the HPC resources of IDRIS under the allocation 2025-
AD011015117R2 made by GENCI. The authors wish to thank the ACL/ARR reviewers for their insightful comments and suggestions. 



\bibliography{anthology,custom}

\appendix
\section {Appendix}
\begin{table*}[htb!]
\centering
\setlength{\tabcolsep}{2.5pt}
\renewcommand{\arraystretch}{0.85}
\resizebox{\textwidth}{!}{%
  \begin{tabular}{|l|*{16}{r@{\hskip 2pt}|}}
  
    \toprule
        & \multicolumn{1}{|c|}{\textbf{GPT-2 }} & \multicolumn{1}{|c|}{\textbf{BLOOM}} & \multicolumn{1}{|c|}{\textbf{\llama-3}} & \multicolumn{1}{|c|}{\textbf{Qwen-3}}
        & \multicolumn{1}{|c|}{\textbf{GPT-2 }} & \multicolumn{1}{|c|}{\textbf{BLOOM}} & \multicolumn{1}{|c|}{\textbf{\llama-3}} & \multicolumn{1}{|c|}{\textbf{Qwen-3}}
        & \multicolumn{1}{|c|}{\textbf{GPT-2 }} & \multicolumn{1}{|c|}{\textbf{BLOOM}} & \multicolumn{1}{|c|}{\textbf{\llama-3}} & \multicolumn{1}{|c|}{\textbf{Qwen-3}} 
        & \multicolumn{1}{|c|}{\textbf{GPT-2 }} & \multicolumn{1}{|c|}{\textbf{BLOOM}} & \multicolumn{1}{|c|}{\textbf{\llama-3}} & \multicolumn{1}{|c|}{\textbf{Qwen-3}}
        \\
    \toprule
        & \multicolumn{4}{c|}{English (1.32m)}
        & \multicolumn{4}{c|}{French (0.97m)}
        & \multicolumn{4}{c|}{German (0.30m)}
        & \multicolumn{4}{c|}{Spanish (1.18m)} \\
    \midrule

    Full vocab
        & 0.66/09.9 & 0.66/5.9 & 0.66/4.5 & 0.69/3.87
        & 0.62/14.9 & 0.63/8.5 & 0.63/6.7 & 0.66/4.44
        & 0.66/20.8 & 0.69/17.6 & 0.71/16.1 & 0.70/5.23
        & 0.63/12.6 & 0.64/6.8 & 0.63/5.2 & 0.65/3.98 \\

    First$_k$
        & 0.62/13.2 & 0.61/7.7 & 0.62/5.1 & 0.63/4.31
        & 0.56/17.9 & 0.59/10.6 & 0.57/8.4 & 0.61/5.42
        & 0.59/29.7 & 0.61/24.1 & 0.61/20.3 & 0.67/7.32/
        & 0.58/16.1 & 0.60/8.6 & 0.58/6.3 & 0.61/4.86 \\

    First$_{k>0}$
        & 0.63/11.3 & 0.63/7.0 & 0.64/5.0 & 0.65/4.03
        & 0.59/16.1 & 0.61/9.9 & 0.61/7.8 & 0.62/5.14
        & 0.65/23.6 & 0.64/23.5 & 0.67/20.0 & 0.68/6.92
        & 0.61/14.2 & 0.61/8.1 & 0.61/6.0 & 0.62/4.56 \\

    Top$_k$
        & 0.63/11.8 & 0.64/6.5 & 0.64/4.8 & 0.66/3.91
        & 0.59/16.1 & 0.61/9.4 & 0.61/7.6 & 0.63/4.85
        & 0.65/23.6 & 0.67/20.3 & 0.67/19.0 & 0.69/6.76
        & 0.61/14.3 & 0.62/7.5 & 0.61/5.4 & 0.63/4.10 \\

    \adaptbpe
        & 0.64/10.1 & 0.64/6.1 & 0.64/4.7 & 0.67/3.90
        & 0.61/15.3 & 0.61/9.0 & 0.61/7.0 & 0.65/4.65
        & 0.65/21.1 & 0.68/18.6 & 0.70/17.3 & 0.70/6.31
        & 0.62/13.0 & 0.62/7.0 & 0.61/5.3 & 0.64/4.02 \\

    \bottomrule
    \toprule

        & \multicolumn{4}{c|}{Occitan (0.34m)}
        & \multicolumn{4}{c|}{Manipuri (0.04m)}
        & \multicolumn{4}{c|}{Swahili (0.38m)}
        & \multicolumn{4}{c|}{Tamil (0.07m)} \\
    \midrule

    Full vocab
        & 0.59/4.8 & 0.60/5.8 & 0.66/4.0 & 0.63/3.42
        & 0.07/2.2 & 0.07/3.6 & 0.08/1.9 & 0.08/3.27
        & 0.60/5.7 & 0.64/6.3 & 0.63/5.3 & 0.67/7.29
        & 0.05/1.9 & 0.06/3.2 & 0.07/2.6 & 0.06/3.85 \\

    First$_k$
        & 0.55/6.0 & 0.56/6.1 & 0.60/5.9 & 0.59/4.56
        & 0.03/3.2 & 0.03/3.7 & 0.03/2.3 & 0.03/3.93
        & 0.56/6.2 & 0.58/6.8 & 0.53/5.8 & 0.62/9.10
        & 0.04/2.2 & 0.05/3.7 & 0.04/2.7 & 0.04/4.26 \\

    First$_{k>0}$
        & 0.56/5.9 & 0.58/6.5 & 0.63/5.3 & 0.60/4.55
        & 0.05/3.3 & 0.05/3.8 & 0.05/2.3 & 0.05/3.87
        & 0.58/6.1 & 0.61/6.8 & 0.61/5.7 & 0.64/8.32
        & 0.05/2.2 & 0.05/3.7 & 0.05/2.7 & 0.05/4.12 \\

    Top$_k$
        & 0.58/5.7 & 0.57/6.1 & 0.58/4.4 & 0.62/4.12
        & 0.03/3.3 & 0.03/3.5 & 0.03/2.1 & 0.05/3.56
        & 0.59/5.9 & 0.61/6.5 & 0.61/5.6 & 0.65/8.20
        & 0.05/2.1 & 0.05/3.4 & 0.05/2.6 & 0.05/4.01 \\

    \adaptbpe
        & 0.58/5.0 & 0.59/6.0 & 0.65/4.2 & 0.62/3.73
        & 0.06/2.3 & 0.06/3.7 & 0.07/2.0 & 0.06/3.34
        & 0.60/5.8 & 0.62/6.4 & 0.62/5.5 & 0.66/8.03
        & 0.05/2.0 & 0.05/3.4 & 0.06/2.6 & 0.06/3.85 \\

    \bottomrule
  \end{tabular}
}
\caption{Compression utility and perplexity for $N=15$k merges on the Wikipedia test sets. Each cell reports utility/perplexity.}
\label{tab:monolingual-full}
\end{table*}

\subsection{\adaptbpe{} at work \label{ssec:illustration}}
\begin{figure*}
    \centering
    \includegraphics[width=\textwidth]{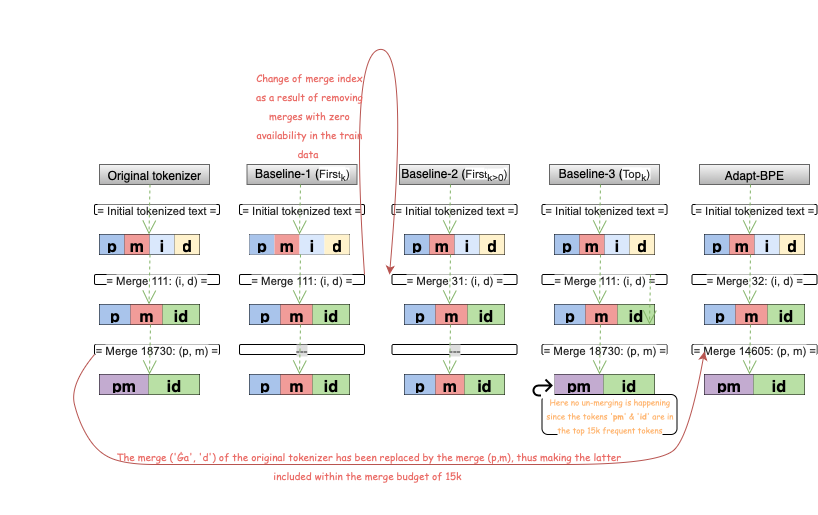}
    \caption{An illustration of BPE merging unmerging procedures The selected token is \textbf{"pmid"} extracted from PubMed data and the original tokenizer is BLOOM.}
    \label{fig:ann:illustration}
\end{figure*}

\begin{figure}[t]
    \centering
    \includegraphics[width=1.0\linewidth]{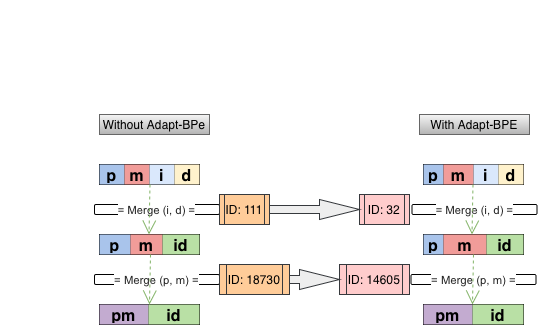}
    \caption{An illustration of merge order change with \adaptbpe{}. The selected token is \textbf{"pmid"} extracted from PubMed data and the original tokenizer is BLOOM}
    \label{fig:ann:illustration_order}
\end{figure}

Figure~\ref{fig:ann:illustration} illustrates the merge / unmerge procedure for various pruning strategies and Figure~\ref{fig:ann:illustration_order} exhibit the merge order change with and without \adaptbpe{}.

\subsection{Monolingual Evaluation}
\label{ssec:app:cupp}
We report in Table~\ref{tab:monolingual-full} the compression utility and perplexity for the full set of eight languages considered in our experiments, also reporting results obtained with Qwen-3 \citep{bai-2023-qwen}. For Qwen, our experiments use the 8 billion parameter model; the associated tokenizer contains 151,656 pieces.

\subsection{Text classification: SIB \label{ssec:app:sib}}
\done\todo{Complete this}
We fine tune the \llama{} model with 701 training instances as follows. We first replace the expected category labels in the training data with numbers (from 1 to 7), so that the model outputs are not affected by changes in tokenization. Fine-tuning is performed with LoRA \cite{hu-2022-lora} implemented in the HuggingFace library\footnote{\url{https://huggingface.co/docs/peft/main/en/conceptual_guides/lora}}, with training parameters of 3 epochs, 2e-4 learning rate. In inference, we generate 5 tokens with greedy decoding, and match the expected label with the first token.

\subsection{Computational Costs \label{ssec:ann:compute}}

Trimming the \llama{} vocabulary down to 15k units enables a potential saving of about 900m parameters in the embedding and the final layers when generating texts; as the projection matrix needs to be stored in memory, pruning could also significantly impact the memory total footprint. For the experiments with Machine Translation (EMEA, English into French) of section~\ref{ssec:machtrans}, we monitored the cost-benefit tradeoff of using \adaptbpe{} compared to the full model: the number of tokens, consequently the average sentence length, increases by 3.6\%; yet, the total computation time for 1k sentences is reduced by approximately 2 minutes (out of 23.9, a reduction of 8.4\%), owing to faster computations of the softmax in the output layer. Even without making any change to the inference code, or trying to optimize the vocabulary size in inference, we still see a clear gain in using \adaptbpe{} over the full model. Details are in Table~\ref{tab:inference-time}.

\begin{table*}[!t]
\centering
\resizebox{\columnwidth}{!}{%
\begin{tabular}{lccc}
\toprule
\textbf{Tokenizer} & \textbf{Time (s)} & \textbf{Tokens} & \textbf{Tok/s} \\
\midrule
Original   & 1548.10 & 30,804 & 19.90 \\
\adaptbpe{} & 1432.88 & 31,904 & 22.27 \\
\bottomrule
\end{tabular}}
\caption{Inference times for \llama{} computing the translation of 1k EMEA sentence pairs, comparing the original and \adaptbpe{} tokenizers.}
\label{tab:inference-time}
\end{table*}

\end{document}